\documentclass{article}

\PassOptionsToPackage{numbers, compress}{natbib}
\usepackage[final]{neurips_2023}

\usepackage[utf8]{inputenc} % allow utf-8 input
\usepackage[T1]{fontenc}    % use 8-bit T1 fonts
\usepackage{hyperref}       % hyperlinks
\usepackage{url}            % simple URL typesetting
\usepackage{booktabs}       % professional-quality tables
\usepackage{amsfonts}       % blackboard math symbols
\usepackage{nicefrac}       % compact symbols for 1/2, etc.
\usepackage{microtype}      % microtypography
\usepackage{xcolor}         % colors

\usepackage{amsmath,amsfonts,bm}

\def\eqref#1{equation~\ref{#1}}
\def\1{\bm{1}}

\DeclareMathAlphabet{\mathsfit}{\encodingdefault}{\sfdefault}{m}{sl}
\SetMathAlphabet{\mathsfit}{bold}{\encodingdefault}{\sfdefault}{bx}{n}

\usepackage{subcaption}
\usepackage{caption}
\usepackage{array,graphicx,float}
\usepackage{soul} % for strike-through \st command
\usepackage{tabularray}
\usepackage{wrapfig}
\usepackage{multirow}

\title{Leveraging Diffusion Disentangled Representations to Mitigate Shortcuts in Underspecified Visual Tasks}

\author{Luca Scimeca$^1$, Alexander Rubinstein$^2$, \\
\textbf{Armand Mihai Nicolicioiu$^3$, Damien Teney$^3$ \& Yoshua Bengio$^{1,4}$} \vspace{5pt}\\
$^1$ Mila - Quebec AI Institute and Université de Montréal, Quebec, Canada \\
$^2$ Eberhard-Karls-Universität Tübingen, Germany \\
$^3$ Idiap Research Institute, Switzerland \\
$^4$ CIFAR Senior Fellow \\
\texttt{luca.scimeca@mila.quebec} \\
}

\begin{document}

\maketitle

\vspace{-10pt}
\begin{abstract}
Spurious correlations in the data, where multiple cues are predictive of the target labels, often lead to shortcut learning phenomena, where a model may rely on erroneous, easy-to-learn, cues while ignoring reliable ones. In this work, we propose an ensemble diversification framework exploiting the generation of synthetic counterfactuals using Diffusion Probabilistic Models (DPMs). We discover that DPMs have the inherent capability to represent multiple visual cues independently, even when they are largely correlated in the training data. We leverage this characteristic to encourage model diversity and empirically show the efficacy of the approach with respect to several diversification objectives. We show that diffusion-guided diversification can lead models to avert attention from shortcut cues, achieving ensemble diversity performance comparable to previous methods requiring additional data collection.
\end{abstract}

\section{Introduction} \label{sec:intro}

% PROBLEM STATEMENT
Deep Neural Networks (DNNs) have achieved unparalleled success in countless tasks across diverse domains. However, they are not devoid of pitfalls. 
% ---- What is shortcut learning
One such downside manifests in the form of shortcut learning, a phenomenon whereby models latch onto simple, non-essential cues that are spuriously correlated with target labels in the training data~\cite{Geirhos2020, shah2020pitfalls, scimeca2022shortcut}. Often engendered by the under-specification present in data, this simplicity bias often presents easy and viable \emph{shortcuts} to performing accurate prediction at training time, regardless of a model's alignment to the downstream task. For instance, previous work has found models to incorrectly rely on background signals for object prediction \cite{xiao2021noise, Beery_2018_ECCV}, or to rely on non-clinically relevant metal tokens to predict patient conditions from X-Ray images~\cite{Zech2018-qx}.
% --- why is it a problem
Leveraging shortcut cues can be harmful when deploying models in sensitive settings. For example, shortcuts can lead to the reinforcement of harmful biases when they endorse the use of sensitive attributes such as gender or skin color~\cite{EstimatingAndMitigating2019, InvestigatingBias2020, scimeca2022shortcut}.

To overcome the simplicity bias, various methods have strived to find ways to drive models to use a diversified set of predictive signals. In ensemble settings, for example, a collection of models has been trained while ensuring mutual orthogonality of their input gradients, effectively driving models to attend to different locations of the input space for prediction~\cite{ross2020ensembles, teney2022evading, teney2022predicting, nicolicioiu2023learning}.
%, effectively constructing a diverse set of hypotheses by encouraging each dimension of a shared input to have a different effect on the prediction. 
These input-centric methods, however, may be at a disadvantage in cases where different features must attend to the same area of the input space. A more direct approach has instead been via the diversification of the models' outputs. This approach hinges on the availability of additional data that is, at least in part, free of the same shortcuts as the original training data. Through a diversification objective, the models are then made to \emph{disagree} on these samples, effectively fitting functions with different extrapolation tendencies \cite{lee2022diversify, pagliardini2022agree}. This dependency on Out-Of-Distribution (OOD) auxiliary data, however, poses limitations, as this is often not readily accessible, and can be costly to procure. %, and must, at least in part, be free of the same shortcuts as the original training data.

% GOAL/constraints
The primary objective of this work is to mitigate shortcut learning tendencies, particularly when they result in strong, unwarranted biases, access to additional data is expensive, and different features may rely on similar areas of the input space. To achieve this objective, we propose an ensemble diversification framework relying on the generation of synthetic counterfactuals for model disagreement. These synthetic samples should: first, lie in the manifold of the data of interest; and second, at least be in part free of the same shortcuts as the original training data. We make use of Diffusion Probabilistic Models (DPMs) to generate the counterfactuals for ensemble disagreement.

\begin{figure}
\centering
\vspace{-20pt}
\includegraphics[width=.9\linewidth]{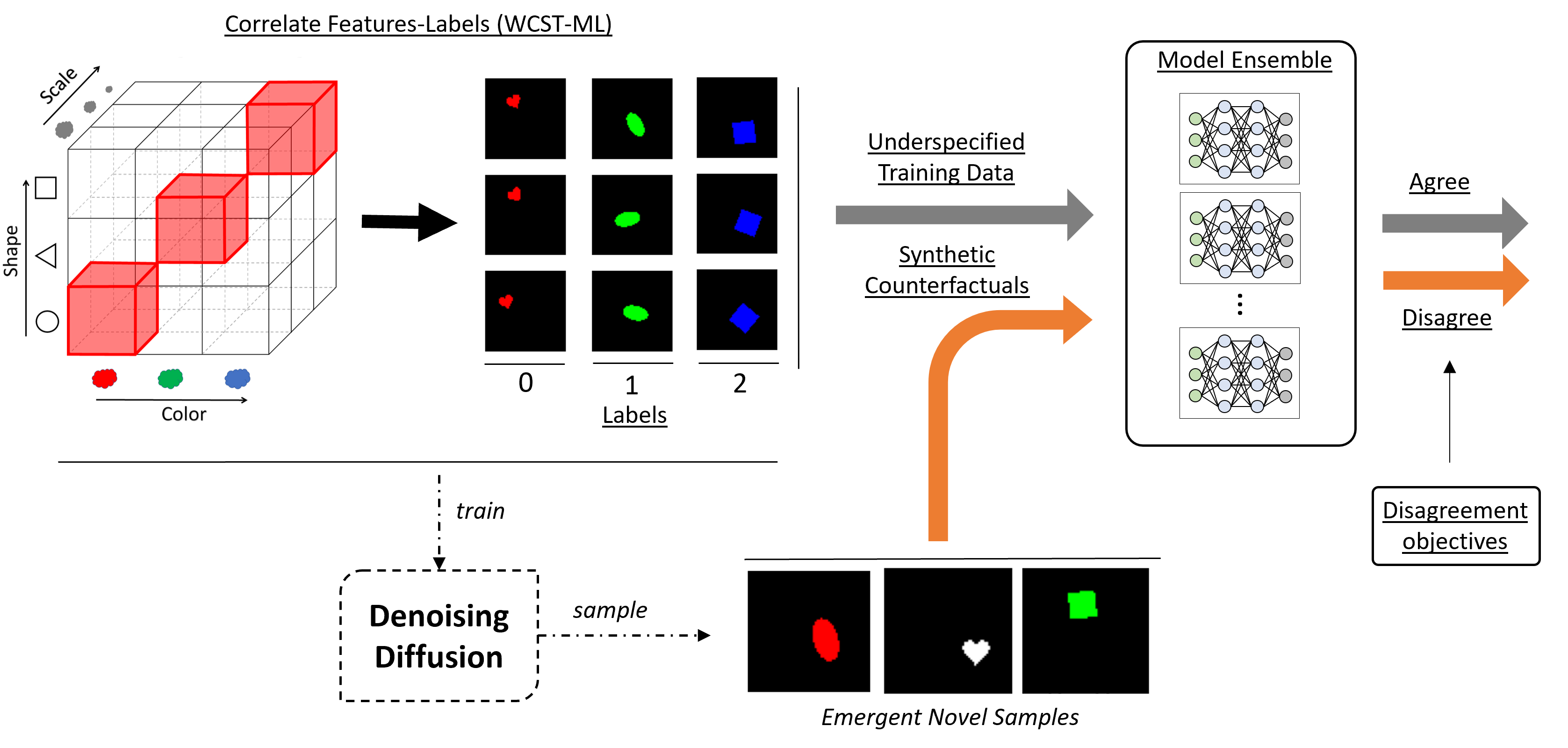}
\caption{Experimental Framework.} 
\label{fig:framework}
\vspace{-10pt}
\end{figure}

% diff models and hypothesis
In recent years, Diffusion Probabilistic Models have emerged as a transformative generative tool. Numerous studies have underscored their prowess in generating synthetic images, which can then be harnessed to enrich datasets and bolster classification performance~\cite{sariyildiz2023fake, azizi2023synthetic, yuan2022not}. 
In several cases, DPMs have been shown to transcend surface-level statistics of the data, making them invaluable in understanding data distributions and features \cite{chen2023beyond, yuan2022not, wu2023uncovering}. Most importantly, recent studies have indicated the ability of DPMs to achieve feature disentanglement via denoising reconstructions~\cite{kwon2022diffusion, wu2023uncovering}. 

% plan 
We first aim to test whether DPMs can achieve feature disentanglement even when training on samples showcasing fully correlated features. Hinging upon this crucial characteristic, we hypothesize that diversification, and shortcut mitigation, can be achieved via ensemble disagreement on the samples generated by the DPMs. These samples provide models with an opportunity to break the spurious correlations encouraged during training. In our scenario, we show how sampling from DPMs can lead to an effective mixture of features immediately available to be leveraged for ensemble diversification. Remarkably, without a need to control for the diversity in our synthetic samples, our experiments confirm that the extent and quality of our diffusion-guided ensemble diversification is on par with existing methods that rely on additional data. 

Our contributions are three-fold:
\begin{enumerate}
    \item We show that DPMs can achieve feature disentanglement even when training on samples showcasing fully correlated features.
    \item We demonstrate that diverse ensemble models can achieve shortcut cue mitigation within WCST-ML, a benchmark designed to examine a special case of shortcut learning, where each feature perfectly correlates with the prediction labels.
    \item We show that DPMs can be used to synthesize counterfactuals for ensemble disagreement, leading to state-of-the-art diversification and shortcut bias mitigation.
\end{enumerate}

\section{Results} \label{sec:results}
We leverage two representative datasets, a color-augmented version of DSprites \cite{dsprites17} (ColorDSprites) and UTKFace \cite{zhang2017age}, as the experimental grounds for our study, previously shown to contain features leading to strong preferential cue bias by models \cite{scimeca2022shortcut}. We apply the WCST-ML framework, creating training datasets of fully correlated feature-labels groups (Figure \ref{fig:framework}). For ColorDSprites, we consider the features of $K_{DS}=\{color, orientation, scale, shape\}$. Within UTKFace we consider the features of $K_{UTK}=\{ethnicity, gender, age\}$. See Section \ref{sec:suppl:dataset} for  implementation details.

\subsection*{Diffusion Counterfactuals Display Disentanglement Capabilities}\label{sec:res:diff_disentanglement}

\begin{figure}
\centering
\vspace{-15pt}
\includegraphics[width=.8\linewidth]{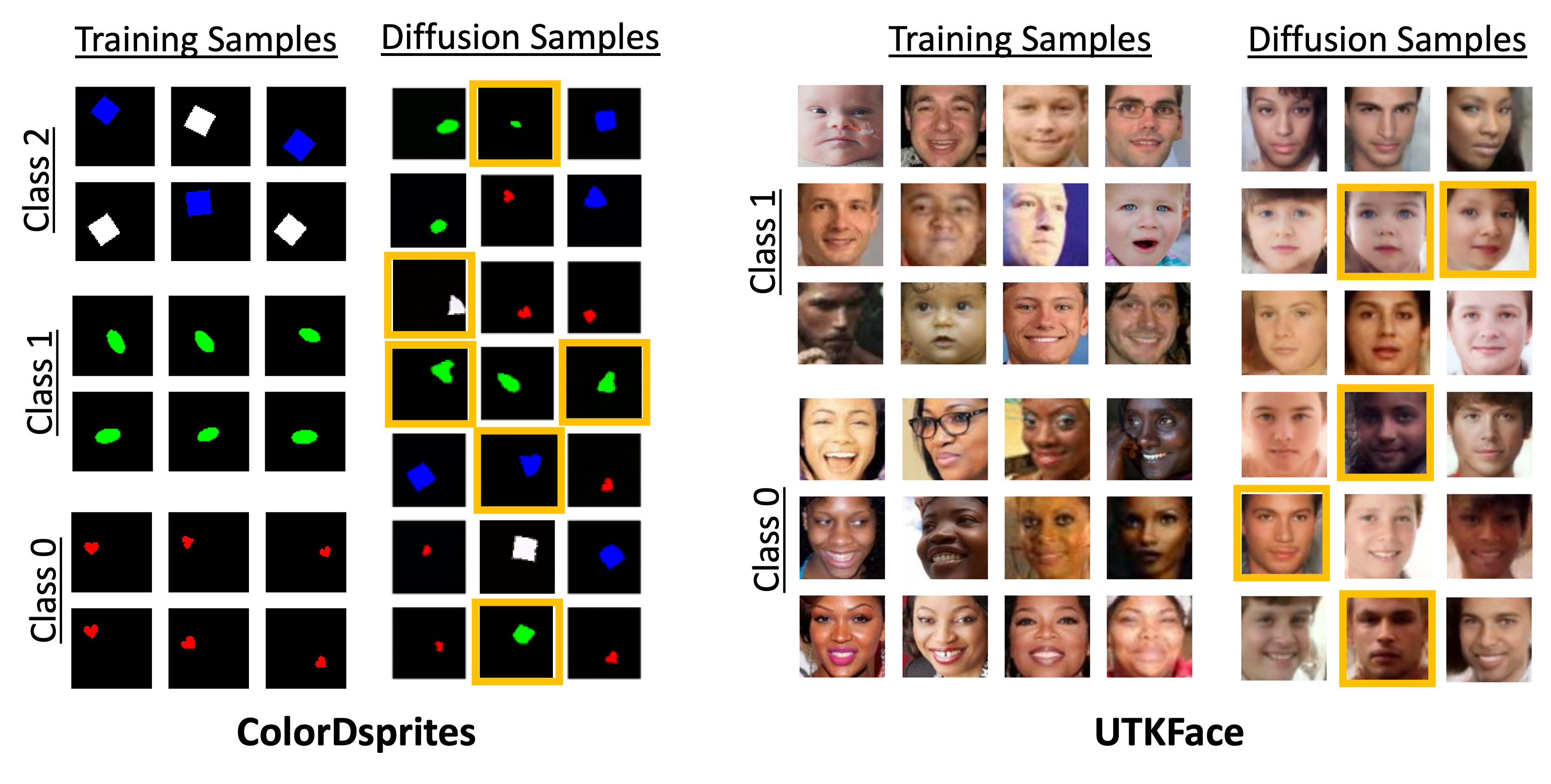}
\caption{Training and Diffusion samples for ColorDSprites and UTKFace based on WCST-ML. 
We observe novel objects with mixtures of correlated features in the diffusion-generated samples.
} 
\label{fig:diff_sampling}
\end{figure}

% fig showing training data, then diffusion sampling (diff. levels of fidelity?)
We test the ability of DPMs to transcend surface-level statistics of the data and achieve feature disentanglement, even in the presence of full correlations. We train the DPMs on both ColorDSprites and UTKFace and show in Figure \ref{fig:diff_sampling} both training samples (left halves) and samples from the trained DPMs (right halves). We observe how sampling from the trained DPMs generates previously unseen feature combinations, despite the fully correlated coupling of features during training (e.g. $\langle green/white, heart \rangle$, $\langle green, square \rangle$ or $\langle child, female \rangle$), displaying innate disentangling capabilities in their latent space. Leveraging this capability is crucial in the generation of samples for disagreement, as it allows models to break from the shortcuts in the training distribution.

\begin{figure}
\centering
\begin{subfigure}[b]{.45\textwidth}
    \includegraphics[width=.98\linewidth]{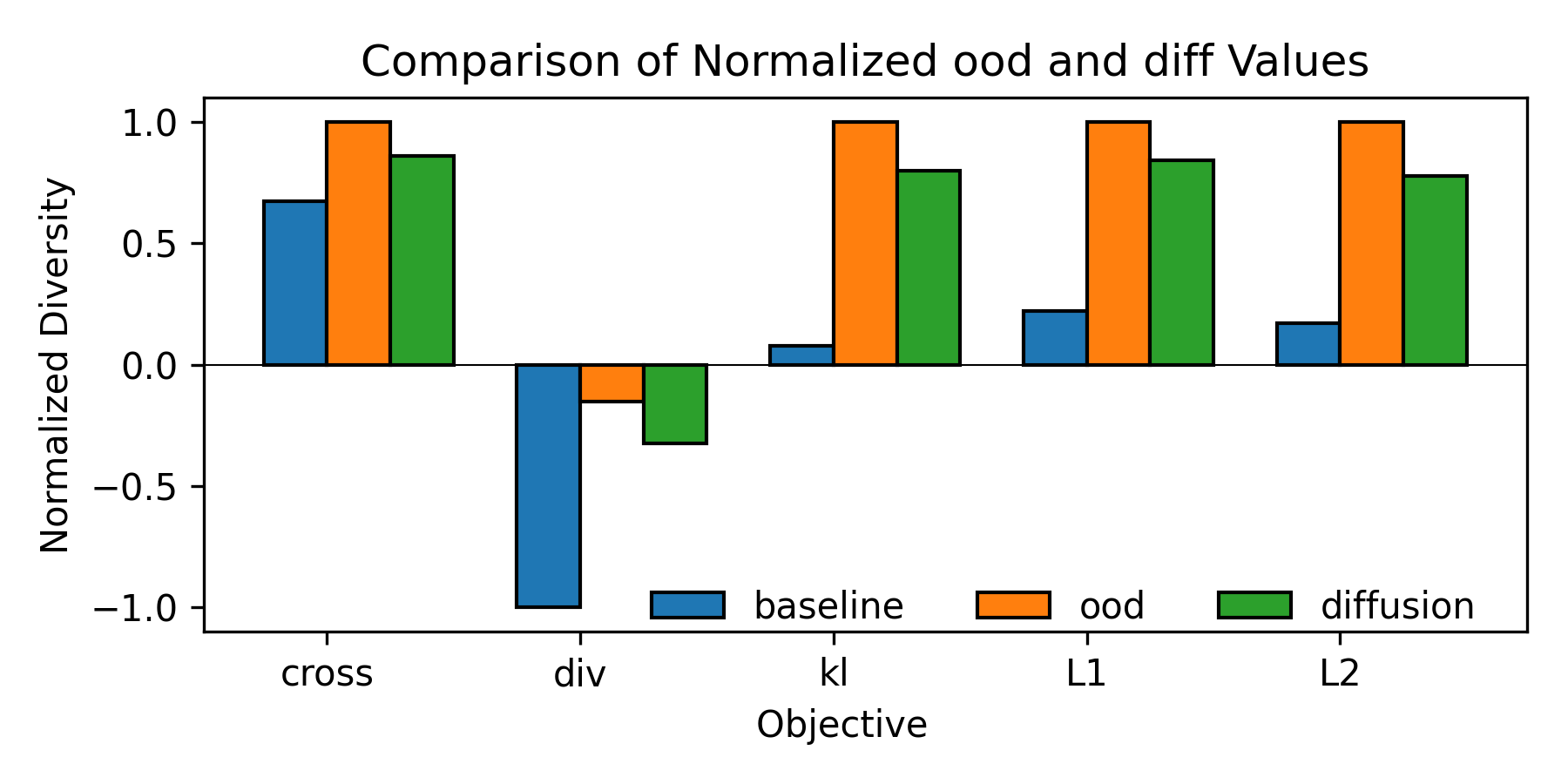}
    \caption{ColorDSprites} 
\end{subfigure}
\begin{subfigure}[b]{.45\textwidth}
    \includegraphics[width=.98\linewidth]{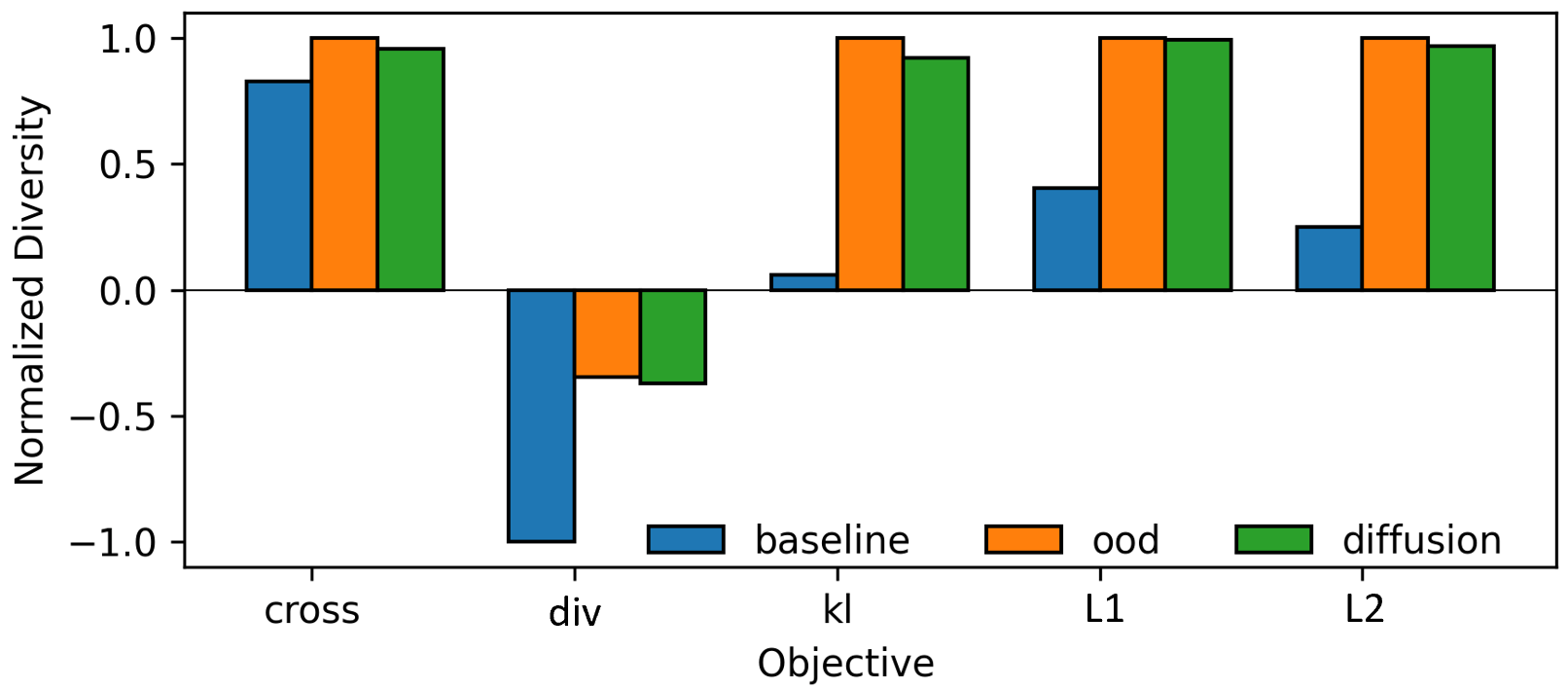}
    \caption{UTKFace} 
\end{subfigure}
\caption{Diversity Comparison across metrics (higher is better).} 
\label{fig:diff_ood_comparison}
\vspace{-10pt}
\end{figure}

\subsubsection*{Diffusion-guided Diversity Leads to Comparably Diverse Ensembles}

We train an ensemble, comprising 100 ResNet-18 models, on both UTKFace and ColorDSprites. The ablation studies comprised 5 diversification objectives, denoted by ${{obj} \in \{\ div,\ cross,\ l1,\ l2,\ kl\}}$. The \emph{div} diversification objective is adapted from \cite{lee2022diversify}; The \emph{cross} objective aims at diversifying the predictions of two models by minimizing their negative mutual cross-entropy; the $kl$ objective aims at maximizing the $kl$ divergence between the output distributions of any two models; the $l1$ and $l2$ objectives are baseline objectives which aim to maximize the pairwise distance between any two model outputs, and the moving average of the ensemble prediction. While optimizing a cross-entropy loss on the training data, each objective is computed on a separate set of samples which we will refer to as `ood' when coming from left-out, feature-uncorrelated, samples from the original dataset, and `diffusion' when generated by a DPM; The `baseline' considers the case where no diversification objective is added. We direct the reader to Suppl. Section \ref{sec:suppl:ensemble_training} for implementation details. 
% For both baseline and diversification experiments, we favor different training dynamics by using separate vanilla Adam optimizers for each model.

In Figure \ref{fig:diff_ood_comparison} we compare the objective-wise normalized diversity achieved in each scenario by the ensemble. We find the diffusion-led diversity to be consistently within 5\% from the metrics achieved when using pure OOD samples, both typically over 50\% higher than the baseline.

\subsection*{Ensemble Diversification to Break Shortcut Learning Biases}

By WCST-ML, we can test each model's bias to a cue by testing the model's output on purposefully designed test sets \cite{scimeca2022shortcut}. We test the quality of the diversification obtained in Table \ref{tab:diversity:all}, and ascertained the fraction of models attending to specific cues when trained on OOD and diffusion-generated samples respectively. A model was deemed to be attending to a cue if its validation accuracy on the cue-specific out-of-distribution (OOD) WCST-ML dataset was highest relative to all other features. Our baseline findings mirrored the observations in \cite{scimeca2022shortcut}. Specifically, each model in the baseline ensemble predominantly attended the \emph{color} cue in ColorDSprites and the \emph{ethnicity} cue in UTKFace, showcasing strong preferential cue bias while achieving near-perfect classification of in-distribution validation samples. Notably, upon introducing the diversification objectives, we observed a perceptible shift in the models' behavior, some of which averted their focus from the primary, easy-to-learn cues, turning instead to other latent cues present within the data. Among the objectives considered, $kl$, $l1$, and $l2$ exhibited the highest cue diversity, catalyzing the ensemble to distribute attention across multiple cues. However, this also came at the expense of a marked drop in the average ensemble performance. Conversely, the \emph{div} and \emph{cross} objectives yielded milder diversification, focusing primarily on the next readily discernible cues, specifically \emph{scale} for ColorDSprites and \emph{gender} for UTKFace, all the while maintaining a generally higher ensemble validation performance. As confirmed in Figure \ref{fig:diff_ood_comparison}, the diversification in Table \ref{tab:diversity:all} is largely comparable, with approximately an average of $10\%$ to $30\%$ of the models averting their attention from the main cue in ColorDSprites, and up to $40\%$ of the models averting their attention to the ethnicity cue in UTKFace for both.

\begin{table} % table* will make the table span both columns
\centering
\vspace{-15pt}
\caption{Results on ColorDSprites and UTKFace when using real OOD samples(upper tables), or Diffusion-generated counterfactuals (lower tables), for model disagreement. The feature columns report the fraction of models biased towards the respective features. The final column reports the average validation accuracy for the ensemble on a left-out feature-correlated \emph{diagonal} set, of the same distribution as the original training data.}
\label{tab:diversity:all}

\begin{subtable}{.99\textwidth} % table* will make the table span both columns
\centering
\small
\caption*{OOD Disagreement}
\vspace{-10pt}
\label{tab:diversity_ood}

\begin{minipage}{0.53\textwidth}
\centering
\small
\caption*{ColorDSprites}

\resizebox{\linewidth}{!}{%

\begin{tabular}{c|cccc|c}
                  & \textbf{color} & \textbf{orientation} & \textbf{scale} & \textbf{shape} & \begin{tabular}[c]{@{}c@{}}\textbf{valid. accuracy }\\(mean +/- std)\end{tabular}  \\ 
\hline
\textbf{baseline} & 1.00          & 0.00                & 0.00          & 0.00          & 1.000 +/- 0.00                                                                     \\
\textbf{cross}    & 0.90          & 0.00                & 0.10          & 0.00          & 0.791 +/- 0.19                                                                     \\
\textbf{div}     & 0.91          & 0.00                & 0.06          & 0.03          & 0.744 +/- 0.22                                                                     \\
\textbf{kl}       & 0.68          & 0.05                & 0.15          & 0.12          & 0.583 +/- 0.30                                                                     \\
\textbf{l1}      & 0.67          & 0.02                & 0.17          & 0.14          & 0.621 +/- 0.24                                                                     \\
\textbf{l2}      & 0.78          & 0.02                & 0.14          & 0.06          & 0.683 +/- 0.24                                                                    
\end{tabular}
}

\end{minipage}
\hfill
\begin{minipage}{0.43\textwidth}
\centering
\small
\caption*{UTKFace}
\resizebox{\linewidth}{!}{%

\begin{tabular}{c|ccc|c}
\textbf{ }        & \textbf{age} & \textbf{ethnicity} & \textbf{gender} & \begin{tabular}[c]{@{}c@{}}\textbf{valid. accuracy} \\(mean +/- std)\end{tabular}  \\ 
\hline
\textbf{baseline} & 0.00        & 1.00              & 0.00           & 0.931 +/- 0.02                                                                     \\
\textbf{cross}    & 0.00        & 0.81              & 0.19           & 0.904 +/- 0.03                                                                     \\
\textbf{div}      & 0.00        & 0.92              & 0.08           & 0.903 +/- 0.02                                                                     \\
\textbf{kl}       & 0.12        & 0.59              & 0.29           & 0.604 +/- 0.24                                                                     \\
\textbf{l1}      & 0.07        & 0.63              & 0.30           & 0.705 +/- 0.15                                                                     \\
\textbf{l2}      & 0.05        & 0.62              & 0.33           & 0.698 +/- 0.16                                                                    
\end{tabular}

}
\end{minipage}
\end{subtable}

\begin{subtable}{.99\textwidth} % table* will make the table span both columns
\centering
\small
\caption*{Diffusion Disagreement}
\vspace{-10pt}
\label{tab:diversity_diff}

\begin{minipage}{0.53\textwidth}
\centering
\small
\caption*{ColorDSprites}

\resizebox{\linewidth}{!}{%

\begin{tabular}{c|cccc|c}
              & \textbf{color} & \textbf{orientation} & \textbf{scale} & \textbf{shape} & \begin{tabular}[c]{@{}c@{}}\textbf{valid. accuracy }\\(mean +/- std)\end{tabular}  \\ 
\hline
\textbf{baseline} & 1.00          & 0.00                & 0.00          & 0.00          & 1.000 +/- 0.00                                                                     \\
\textbf{cross}    & 0.89          & 0.00                & 0.10          & 0.01          & 0.803 +/- 0.19                                                                     \\
\textbf{div}      & 0.94          & 0.00                & 0.06          & 0.00          & 0.904 +/- 0.17                                                                     \\
\textbf{kl}       & 0.75          & 0.06                & 0.15          & 0.04          & 0.618 +/- 0.24                                                                     \\
\textbf{l1}      & 0.79          & 0.03                & 0.15          & 0.03          & 0.676 +/- 0.23                                                                     \\
\textbf{l2}      & 0.78          & 0.03                & 0.14          & 0.05          & 0.695 +/- 0.22                                                                    
\end{tabular}
}

\end{minipage}
\hfill
\begin{minipage}{0.43\textwidth}
\centering
\small
\caption*{UTKFace}
\resizebox{\linewidth}{!}{%

\begin{tabular}{c|ccc|c}
              & \textbf{age} & \textbf{ethnicity} & \textbf{gender} & \begin{tabular}[c]{@{}c@{}}\textbf{valid. accuracy} \\(mean +/- std)\end{tabular}  \\ 
\hline
\textbf{baseline} & 0.00        & 1.00              & 0.00           & 0.931 +/- 0.02                                                                     \\
\textbf{cross}    & 0.00        & 0.99              & 0.01           & 0.897 +/- 0.03                                                                     \\
\textbf{div}      & 0.00        & 0.98              & 0.02           & 0.886 +/- 0.03                                                                     \\
\textbf{kl}       & 0.25        & 0.54              & 0.21           & 0.588 +/- 0.26                                                                     \\
\textbf{l1}      & 0.05        & 0.62              & 0.33           & 0.655 +/- 0.16                                                                     \\
\textbf{l2}      & 0.10        & 0.59              & 0.31           & 0.659 +/- 0.17                                                                    
\end{tabular}

}
\end{minipage}
\end{subtable}

\end{table}

% Authors may wish to optionally include extra information (complete proofs, additional experiments and plots) in the appendix. All such materials should be part of the supplemental material after the references in the main text.

\section{Conclusion}
Shortcut learning is a phenomenon undermining the performance and utility of deep learning models, where easy-to-learn cues are preferentially learned for prediction, regardless of their relevance to the downstream task. We investigate the potential of DPMs to aid in the mitigation of this phenomenon by training a diverse ensemble through disagreement on synthetic counterfactuals. We find DPMs able to disentangle the visual feature space of images in two representative datasets, ColorDSprites and UTKFace, and generate novel feature combinations even when trained on images displaying fully-correlated features. We leverage this characteristic for ensemble diversification. Our approach demonstrates negligible performance loss compared to using expensive additional OOD data. Importantly, it competitively averts shortcut cue biases in several ensemble members.

\begin{ack}
Luca Scimeca acknowledges funding from Recursion Pharmaceuticals. Damien Teney is partially supported by an Amazon Research Award.

\end{ack}
\medskip

\bibliography{main}
%%%%%%%%%%%%%%%%%%%%%%%%%%%%%%%%%%%%%%%%%%%%%%%%%%%%%%%%%%%%

\appendix
\section{Appendix}

\subsection{WCST-ML} \label{sec:suppl:wcst-ml}
% WCST-ML, what it is, how it decides train test splits, how it can provide a good test-bed for shortcut-cue learning
To isolate and investigate shortcut bias tendencies, we employ WCST-ML, a method proposed in prior research to dissect the shortcut learning behaviors of deep neural networks~\cite{scimeca2022shortcut}. WCST-ML provides a systematic approach to creating datasets with multiple, equally valid cues, designed to correlate with the target labels. Specifically, given $K$ cues $i_1, i_2, \ldots, i_K$, the method produces a \emph{diagonal} dataset $\mathcal{D}_\text{diag}$ where each cue $i_k$ is equally useful for predicting the labels $Y$. This level playing field is instrumental in removing the influence of feature dominance and spurious correlations, thereby allowing us to observe a model's preference for certain cues under controlled conditions. To rigorously test these preferences, WCST-ML employs the notion of \emph{off-diagonal} samples. These are samples where the cues are not in a one-to-one correspondence with the labels, but instead align with only one of the features under inspection. By evaluating a model's performance on off-diagonal samples, according to each feature, we can test and achieve an unbiased estimate of its reliance on the same. 
% Essentially, high unbiased accuracy for a cue $k$, denoted as $\text{acc}_k(f)$, signals that the model $f$ is heavily relying on that cue for its predictions.

\subsection{Datasets} \label{sec:suppl:dataset}

\paragraph{DSprites:} DSprites comprises a comprehensive set of symbolic objects generated with variations in five latent variables: shape, scale, orientation, and X-Y position. Additionally, we augment this dataset with a color dimension, resulting in \(2,949,120\) distinct samples, we refer to this as ColorDSprites. Within the context of WCST-ML, ColorDSprites permits the examination of shortcut biases in a highly controlled setup. 

\paragraph{UTKFace:} UTKFace provides a dataset of \(33,488\) facial images annotated with attributes like age, gender, and ethnicity. Unlike DSprites, UTKFace presents a real-world, less controlled setup to study bias. Its inherent complexity and diversity make it an ideal candidate for understanding the model's cue preferences when societal and ethical concerns are at stake. 

\subsubsection{Operationalizing WCST-ML across Datasets}
We follow the set-up in~\cite{scimeca2022shortcut} and construct a balanced dataset $\mathcal{D}_\text{diag}$, which includes a balanced distribution of cues, coupled with their corresponding off-diagonal test sets (one for each feature). 
For both datasets, we define a balanced number of classes \(L\) for each feature under investigation. Where the feature classes exceed \(L\), we randomly sub-sample to maintain uniformity. For instance, for the continuous feature `age' in UTKFace, we dynamically select age intervals to ensure the same \(L\) number of categories as other classes, as well as sample balance within each category. For ColorDSprites, we consider the features of $K_{DS}=\{color, orientation, scale, shape\}$ and $L=3$ as constrained by the number of shapes in the dataset. Within UTKFace we consider the features of $K_{UTK}=\{ethnicity, gender, age\}$ and $L=2$ as constrained by the binary classification on \emph{gender}. For each feature set within the respective datasets, we create one dataset of fully correlated features and labels, available at training time, and $K_{DS}$ and $K_{UTK}$ feature-specific datasets to serve for testing the models' shortcut bias tendencies.

\subsection{Ensemble Training and Diversification} \label{sec:suppl:ensemble_training}

In our setup, we wish to train a set of models within an ensemble while encouraging model diversity. Let $f_i$ denote the i$^{th}$ model predictions within an ensemble consisting of $N_m$ models. Each model is trained on a joint objective comprising the conventional cross-entropy loss with the target labels, complemented by a diversification term, represented as ${L}_{div}$. The composite training objective is then:

\begin{equation}
\mathcal{L} = \mathcal{L}_{xent} + \mathcal{L}^{obj}_{div}
\end{equation}

Where $\mathcal{L}_{xent}$ is the cross-entropy loss and $\mathcal{L}^{obj}_{div}$ is the diversification term for a particular objective. To impart diversity to the ensemble, we investigated five diversification objectives denoted by ${{obj} \in \{\ div,\ cross,\ l1,\ l2,\ kl\}}$. The \emph{div} diversification objective is adapted from \cite{lee2022diversify}, and can be summarized as:

\begin{equation}
\mathcal{L}^{div}_{reg} = \lambda_1 \sum_{i \neq j} D_{KL}(p(f_i, f_j) || p(f_i) \otimes p(f_j)) + \lambda_2 \sum_i D_{KL}(p(f_i) || p(y))
\end{equation}

The first term of the sum encourages diversity by minimizing the mutual information from any two predictors, and the second acts as a regularizer to prevent degenerate solutions. 

% The strengths of these components are controlled by hyperparameters $\lambda_1$ and $\lambda_2$, set to 1 throughout the experiments.
% KL OBJECTIVE
We consider other baseline objectives for comparison. The \emph{cross} objective aims at diversifying the predictions of two models by minimizing their negative mutual cross-entropy ($CE$):

\begin{equation}
\mathcal{L}^{cross}_{reg} = -\lambda \sum_{i \neq j} \frac{1}{2} (CE(f_i, \text{argmax}(f_j)) + CE(\text{argmax}(f_i), f_j))
\end{equation}

The \emph{kl} diversification objective aims at maximizing the $kl$ divergence between the output distributions of any two models:

\begin{equation}
    L^{kl}_{reg} = - \lambda \sum_{i \neq j} D_{KL}(f_i || f_j)
\end{equation}

% L1 L2 OBJECTIVES
For the \emph{l1} and \emph{l2} objectives, our goal was to maximize the pairwise distance between any two model outputs and the moving average of the ensemble prediction, thus:

\begin{align}
    L^{l1}_{reg} &= -\lambda \frac{1}{N_m} \sum_{i=1}^{N_m}  \mid f_i - \frac{1}{N_m} \sum_{j=1}^{N_m} f_j  \mid \\
    L^{l2}_{reg} &= -\lambda \frac{1}{N_m} \sum_{i=1}^{N_m} ( f_i - \frac{1}{N_m} \sum_{j=1}^{N_m} f_j ) ^2\\
\end{align}

\subsection{Diffusion Probabilistic Models and Efficient Sampling} \label{sec:suppl:diff_training}

We utilize DPMs to generate synthetic data for our experiments. DPMs function by iteratively adding or removing noise from an initial data point \( x_0 \) via a stochastic process influenced by a predefined noise schedule. The general forward and reverse diffusion processes can be concisely formulated as:

% \begin{align}
%     \text{Forward process:} \quad x_t &\sim q_t(x_t \mid x_{t-1}), \\
%     \text{Reverse process:} \quad z_{t-1} &\sim p_{t-1}(z_{t-1} \mid z_t), \\
%     \text{Sampling procedure:} \quad x_t &= z_t + \sqrt{\beta_t} \epsilon_t, \quad \epsilon_t \sim \mathcal{N}(0, I),
% \end{align}
\begin{align}
    &\text{Forward process:} \quad x_0 \sim p_0(x), \quad x_t \mid x_{t-1} \sim q_t(x_t \mid x_{t-1}),  \\
    &\text{Reverse process:} \quad z_T \sim p_T(z), \quad z_{t-1} \mid z_t \sim p_{t-1}(z_{t-1} \mid z_t), \\
    &\text{Sampling procedure:} \quad z_T \sim p_T(z), \quad x_t = z_t + \sqrt{\beta_t} \epsilon_t, \quad \epsilon_t \sim \mathcal{N}(0, I),
\end{align}

where \( x_t \) and \( z_t \) are the data and latent variables at time \( t \), and \( \beta_t \) represents the noise level at each time step \( t \). The aim is to learn the reverse transition distributions that can map noisy observations back to the original data. The learning objective is often expressed as a variational bound:

\begin{equation}
    \log p_0(x_0) \geq -\sum_{t=1}^{T} \text{KL}(q_t(x_t \mid x_{t-1}) \| p_t(x_t)) + \mathbb{E}_{q_T(x_T)}[\log p_T(z_T)],
\end{equation}
where \( p_t(x_t) = \mathbb{E}_{p_{t-1}(z_{t-1} \mid z_t)}[\mathcal{N}(x_t; z_{t-1}, \beta_t I)] \).

To perform efficient sampling, we use Denoising Diffusion Implicit Models (DDIM) (\cite{song2020denoising}), a first-order ODE solver for DPMs~\cite{salimans2022progressive, lu2022dpm}, which employs a predictor-corrector scheme to reduce the number of sampling steps necessary.

\section{Supplementary Results} \label{sec:suppl:res}
\subsection{Diversification Leads to Ensemble Models Attending to Different Cues} \label{sec:suppl:res_diversification_comparison}
\begin{figure}
    \caption{Comparison of 10 diversified models when training the ensemble while using (a) feature-uncorrelated OOD data and (b) Diffusion Samples.}
    \label{fig:diversity_comparison:all}
    \begin{subfigure}[b]{\linewidth}
    \centering
        \begin{subfigure}[b]{\linewidth}
            \includegraphics[width=\linewidth]{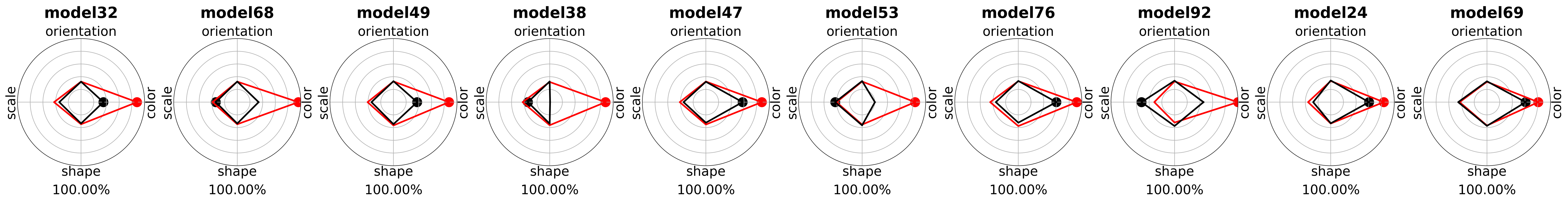}
        \end{subfigure}%
        \hfill
        \begin{subfigure}[b]{.93\linewidth}
            \includegraphics[width=\linewidth]{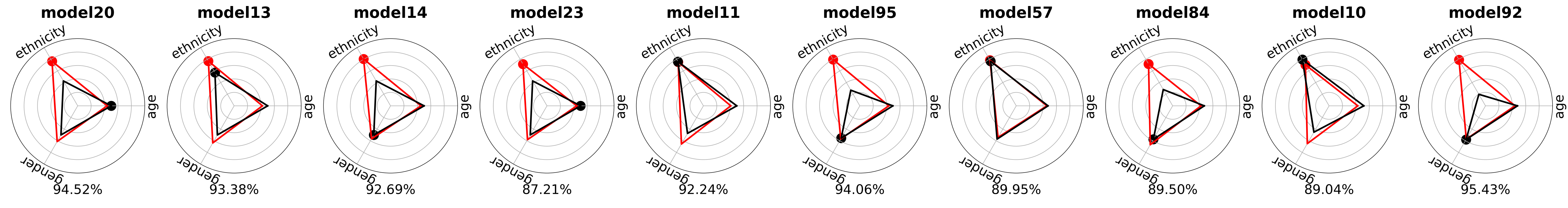}
        \end{subfigure}
    \caption{OOD Disagreement}
    \label{fig:diversity_ood}
    \end{subfigure}

    \begin{subfigure}[b]{\linewidth}
    \centering
        \begin{subfigure}[b]{\linewidth}
            \includegraphics[width=\linewidth]{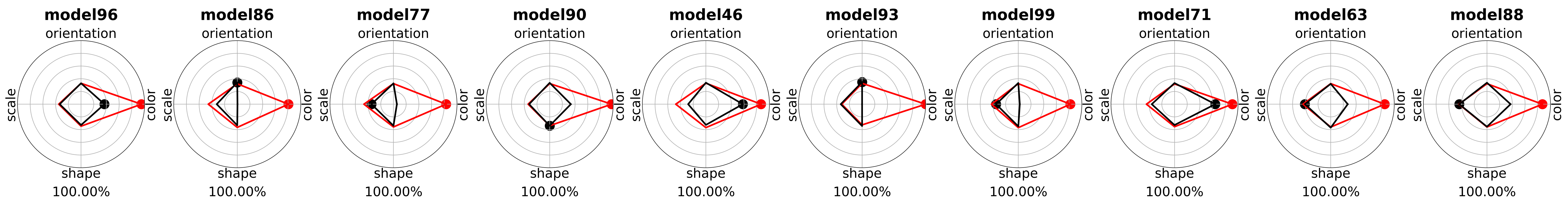}
        \end{subfigure}%
        \hfill
        \begin{subfigure}[b]{.93\linewidth}
            \includegraphics[width=\linewidth]{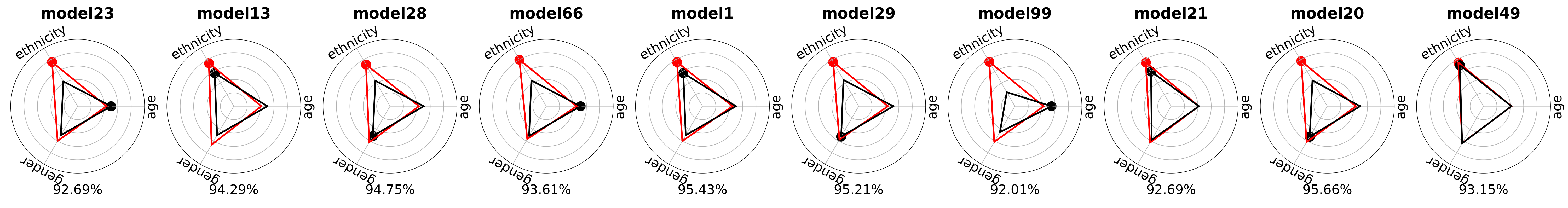}
        \end{subfigure}
    \caption{Diffusion Disagreement}
    \label{fig:diversity_diff}
    \end{subfigure}
\end{figure}

Figure \ref{fig:diversity_comparison:all} illustrates a feature-centric description of 10 ensemble models trained with a diversification objective on OOD data (a) and Diffusion generated counterfactuals (b). Within each results are shown for ColorDSprites (upper panel) and UTKFace (lower panel). The variation across models is evident: several models substantially reduce their dependency on the leading cues of the respective datasets (black edges), diverging considerably from the almost identical configurations present in the baseline ensemble (red edges).

\subsection{On the Influence of an Increased Number of OOD Samples for Ensemble Disagreement} \label{sec:suppl:res_diversification_all_ood}

As per the original objective, with DPM sampling we aim to circumvent the diversification dependency on Out-Of-Distribution data, which is often not readily accessible and can be costly to procure. We test this dependency further and assess the quality of the diversification results when matching the number of OOD data used for diversification to the original training data for the ensemble. We report in Table \ref{tbl:suppl_ood_full} our findings. We observe the quality of the disagreement on ColorDSprites to only marginally benefit from additional disagreement samples, with approximately $10$ to $15\%$ of the models to avert their attention from the shortcut cue \emph{color}. On the other hand, we observe a strong improvement in the diversification for UTKFace, mainly registered via the \emph{cross} objective, where approximately $39\%$ of the models avert their attention from the \emph{ethnicity} shortcut, as opposed to the original $1\%$ to $20\%$ in our previous experiments, while maintaining high predictive performance on the validation set. We speculate this gain to be due to the higher complexity of the features within the data, which may require additional specimens for appropriate diversification.

\begin{table*} % table* will make the table span both columns
\centering
\caption{Diversification results on ColorDSprites and UTKFace when using the same number of OOD samples as the training dataset. The feature columns report the fraction of models (in each row) biased towards the feature. The final column reports the average validation accuracy for the ensemble when tested on a left-out feature-correlated \emph{diagonal} set, of the same distribution as the original training data}

\begin{minipage}{0.53\textwidth}
\centering
\small
\caption*{ColorDSprites}
\label{tbl:suppl_ood_full}

\resizebox{\linewidth}{!}{%
\begin{tabular}{c|cccc|c}
                  & \textbf{color} & \textbf{orientation} & \textbf{scale} & \textbf{shape} & \begin{tabular}[c]{@{}c@{}}\textbf{valid. accuracy }\\\textbf{(mean +/- std)}\end{tabular}  \\ 
\hline
\textbf{baseline} & 1.000          & 0.000                & 0.000          & 0.000          & 1.000 +/- 0.00                                                                              \\
\textbf{cross}    & 0.900          & 0.000                & 0.100          & 0.000          & 0.806 +/- 0.20                                                                              \\
\textbf{div}     & 0.850          & 0.000                & 0.110          & 0.040          & 0.728 +/- 0.22                                                                              \\
\textbf{kl}       & 0.670          & 0.100                & 0.120          & 0.110          & 0.545 +/- 0.27                                                                              \\
\textbf{l1}      & 0.740          & 0.030                & 0.130          & 0.100          & 0.636 +/- 0.27                                                                              \\
\textbf{l2}      & 0.730          & 0.040                & 0.160          & 0.070          & 0.683 +/- 0.26                                                                             
\end{tabular}
}
\end{minipage}
\hfill
\begin{minipage}{0.43\textwidth}
\centering
\small
\caption*{UTKFace}

\resizebox{\linewidth}{!}{%
\begin{tabular}{c|ccc|c}
\textbf{ }        & \textbf{age} & \textbf{ethnicity} & \textbf{gender} & \begin{tabular}[c]{@{}c@{}}\textbf{valid. accuracy} \\(mean +/- std)\end{tabular}  \\ 
\hline
\textbf{baseline} & 0.00        & 1.00              & 0.00           & 0.931 +/- 0.02                                                                              \\
\textbf{cross}    & 0.00        & 0.61              & 0.39           & 0.902 +/- 0.02                                                                              \\
\textbf{div}      & 0.00        & 0.91              & 0.09           & 0.895 +/- 0.02                                                                              \\
\textbf{kl}       & 0.22        & 0.55              & 0.23           & 0.603 +/- 0.24                                                                              \\
\textbf{l1}       & 0.17        & 0.57              & 0.26           & 0.657 +/- 0.18                                                                              \\
\textbf{l2}      & 0.07        & 0.64              & 0.29           & 0.668 +/- 0.17                                                                             
\end{tabular}
}
\end{minipage}
\end{table*}

\end{document}